\documentclass[conference]{IEEEtran}
\IEEEoverridecommandlockouts

\usepackage{amsmath,amssymb,amsfonts}
\usepackage{cite}
\usepackage{graphicx}
\usepackage{textcomp}
\usepackage{xcolor}
\usepackage{bbm}

\usepackage{enumitem}
\usepackage{tabularx}
\usepackage{multirow}
\usepackage{lipsum}
\usepackage{amsmath}
\usepackage{amssymb}
\usepackage{graphicx}
\graphicspath{{./Figures/}}
\usepackage{times}
\usepackage{multicol}
\usepackage{booktabs}
\usepackage{multirow}
\usepackage{makecell}
\usepackage{graphicx}
\usepackage{bm}
\usepackage[nolist]{acronym}
\usepackage{mathtools}
\usepackage{array}
\usepackage{dblfloatfix}
\usepackage{subfigure}
\usepackage{diagbox}
\usepackage{etoolbox}\AtBeginEnvironment{algorithmic}{\small}
\usepackage{soul}
\newif\ifshowcomments
 
 \showcommentstrue  

\ifshowcomments 
    \newcommand{\bae}[1]{\hl{[SB: #1]}\protect\color{black}} 
    \newcommand{\di}[1]{\hl{[DI: #1]}\protect\color{black}} 
    \newcommand{\ft}[1]{\hl{[FT: #1]}\protect\color{black}} 
    \newcommand{\jd}[1]{\hl{[JD: #1]}\protect\color{black}} 
    
\else
    \newcommand{\bae}[1]{}
    \newcommand{\di}[1]{}
    \newcommand{\ft}[1]{}
    \newcommand{\jd}[1]{}
\fi

\usepackage{lipsum}
\usepackage{ragged2e}
\usepackage[colorinlistoftodos]{todonotes}
\usepackage[labelfont=bf, textfont=normalfont, justification=raggedright, singlelinecheck=false]{caption}
\usepackage{algorithm}
\usepackage{algorithmic}

 \usepackage{booktabs}
 
\usepackage{amsthm}

\usepackage{bbm} 
\usepackage[colorlinks = True, linkcolor = blue, citecolor = blue]{hyperref}

\title{\LARGE \bf
VLM-Based Advanced Rider Assistance System for Motorcycle Safety
}

\author{
Mohamed Elnoor$^{1,2}$, Francesca Baldini$^1$, Ananya Trivedi$^{1,3}$,
Faizan M. Tariq$^1$, Jovin D{'}sa$^1$, David Isele$^1$,\\
Sangjae Bae$^1$, Dinesh Manocha$^2$, Yosuke Sakamoto$^1$
\thanks{
All work was done at HRI.
$^1$Honda Research Institute, USA, San Jose, CA, 95134.
$^2$University of Maryland, College Park, MD 20742, USA.
$^3$Northeastern University, Boston, MA 02115, USA.
Corresponding author: \href{mailto:melnoor@umd.edu}{\texttt{melnoor@umd.edu}}.
}
}

\begin{document}

\maketitle

\begin{abstract}
Motorcycles face disproportionately high crash risks compared to cars due to limited protection and heightened sensitivity to surface hazards, yet Advanced Rider Assistance Systems (ARAS) remain underdeveloped relative to  Advanced Driver Assistance Systems (ADAS). We propose a novel ARAS that enhances motorcycle safety through semantic perception and risk-aware planning. Our approach leverages Vision-Language Models (VLMs) for contextual hazard reasoning and integrates them with segmentation-based detection to construct dense risk maps. 
These maps encode both semantic characteristics (e.g., pothole severity, puddle slipperiness) and physical attributes (e.g., size, depth), which produce per-pixel hazard costs that capture motorcycle-specific risks. These maps are used by a sampling-based planner tailored to motorcycle dynamics to recommend throttle and steering actions that minimize hazard exposure while advancing toward the destination. We evaluate our system in different scenarios in the CARLA simulator. Compared to the baseline method, our method achieves higher success rates and lower hazard exposure, while qualitative results demonstrate interpretable risk maps and safe trajectory recommendations. 
\end{abstract}


\section{INTRODUCTION}

Motorcycles and other powered two-wheeled vehicles offer an efficient mode of transportation, particularly in congested urban environments \cite{cox2018environmental}. However, unlike cars, they offer minimal physical protection and are highly sensitive to surface conditions, which makes riders far more vulnerable to road hazards \cite{cheng2011comparison}. Hazards such as potholes, water puddles, uneven surfaces, and debris can significantly impact rider safety, potentially leading to loss of control, skidding, or severe accidents \cite{clarke2004depth}.
This vulnerability is worse in many developing countries, where motorcycle ridership is significantly higher and poor road infrastructure, such as potholes and uneven surfaces, significantly increases crash risk \cite{solagberu2006motorcycle}.  Studies consistently show that motorcycle riders face a crash fatality risk more than twenty times higher than car passengers \cite{ait2024effect}, and human error contributes to nearly 90\% of motorcycle crashes \cite{huth2013predicting}. Common crash scenarios include single-vehicle loss of control on curves, instability caused by surface irregularities, and conflicts at intersections. 
Addressing such challenges, companies such as Honda have announced ambitious goals such as “zero motorcycle fatalities by 2050” \cite{honda2025unrsf},  motivating research on advanced, perception-driven ARAS that proactively reason about and mitigate surface hazards.  Together, these factors highlight the importance of developing ARAS systems that can perform hazard detection and reasoning to support safer motorcycle operation.

\begin{figure}[t]
\centering
\includegraphics[width = \columnwidth]{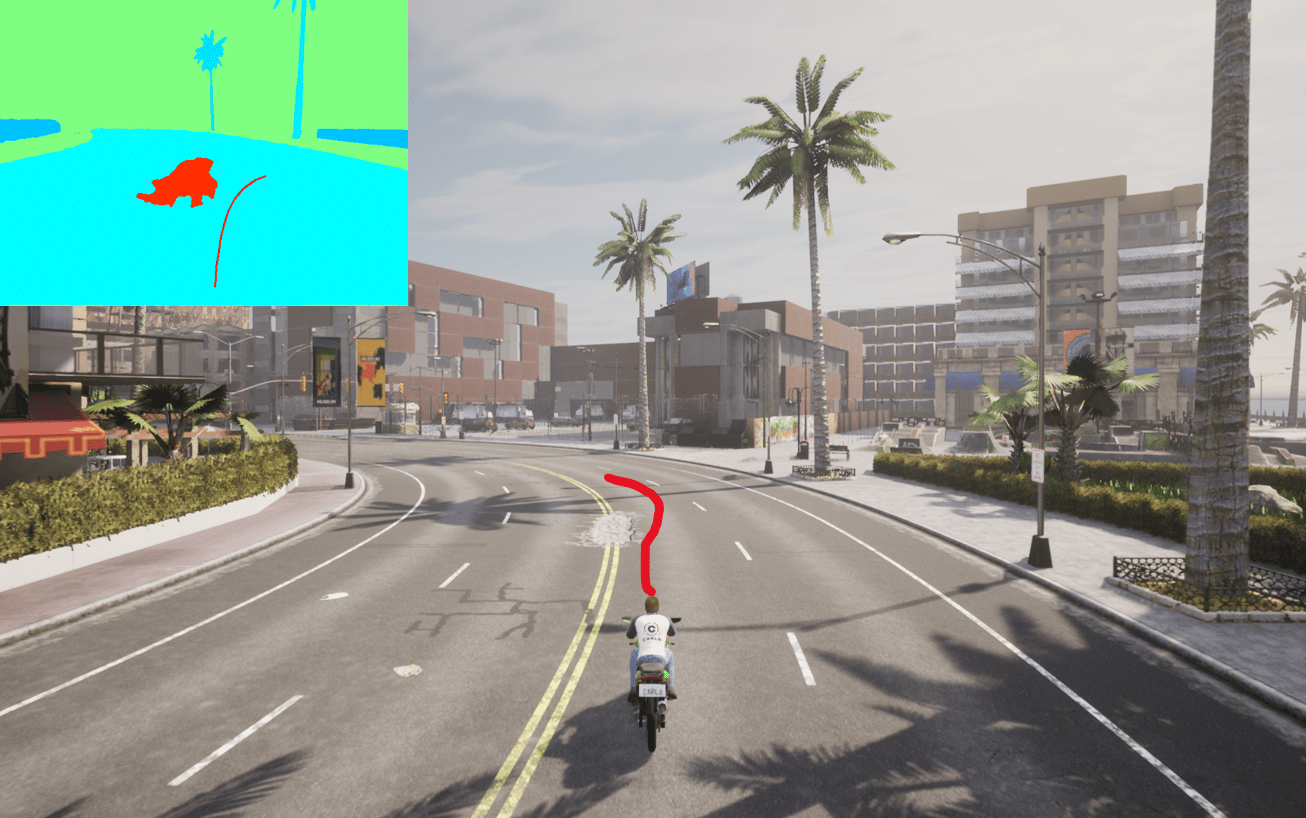} 
\caption{\justifying
A simulated motorcycle rider approaches a pothole in the CARLA simulator. The scene illustrates the risk map on the top left, where our system identifies and localizes the hazard to inform safer motion planning.
}
\label{fig:scenario}
\end{figure}

Early explorations into advanced motorcycle technologies reflect this need. One notable example is the Ghostrider project from the 2005 DARPA Grand Challenges \cite{ieeeGhostrider2005}, which demonstrated one of the first self-balancing, self-navigating motorcycles. Although the platform was capable of short autonomous rides, its limited perception and environmental understanding highlighted the inherent difficulty of achieving robust, context-aware autonomy on two-wheeled vehicles.

Building on these early attempts, ARAS has been proposed as a practical step toward improving motorcycle safety. Research shows that systems such as ABS, autonomous emergency braking, and collision warning could mitigate over 60\% of crashes \cite{ait2024effect}. In contrast, Advanced Driver Assistance Systems (ADAS) for cars, such as lane keeping, forward collision warning, and automatic emergency braking, have matured significantly and demonstrate strong effectiveness in mitigating road hazards \cite{nidamanuri2021progressive}. However, ADAS technologies and their underlying assumptions do not readily transfer to motorcycles, where surface conditions deemed safe for cars (e.g., small potholes, minor unevenness) can pose severe risk to riders due to reduced stability and limited protection.
Despite initiatives such as SAFERIDER \cite{bekiaris2010saferider} and PISa \cite{savino2010pisa}, ARAS development continues to lag behind ADAS. Key challenges include the scarcity of motorcycle-specific datasets, the high cost and safety risks associated with real-world data collection, and the lack of realistic simulation platforms tailored to two-wheeled vehicle dynamics. 

Traditional perception methods, such as semantic segmentation and object detection 
\cite{lai2021mtsan, muhammad2022vision, colley2021effects, murthy2022objectdetect}, 
can identify surface features
like water, asphalt, or potholes, but they treat hazards as static visual categories and do not reason about their implications
for motorcycle stability. These methods cannot assess whether a shallow puddle is safe to ride over, whether a water-filled
pothole poses severe traction loss, or whether a construction sign implies an unseen irregularity beyond the camera’s view.
Consequently, appearance-based perception alone can be insufficient for ARAS, where hazard severity depends on context and motorcycle
dynamics, not category labels alone.

Recent advancements in large Vision-Language Models (VLMs) offer promising ways to address these limitations. VLMs exhibit strong generalization, zero-shot reasoning, and contextual understanding across diverse perception tasks \cite{sural2024contextvlm, chen2024advanced, 11128264}. Unlike conventional models,
they can articulate why a scene element may be dangerous and infer its potential impact on motorcycle safety \cite{xiao2024hazardvlm}. These capabilities align closely with ARAS needs, though real-time deployment remains challenging due to VLMs’ computational cost,
inference latency, and limited spatial understanding.

To address these limitations, we introduce a hybrid ARAS pipeline that combines the semantic reasoning capabilities of VLMs with the spatial precision of segmentation models to construct a dense, hazard-aware risk map for motorcycles. This map representation encodes both the physical attributes and contextual implications of surface hazards,  and is used by a sampling-based planner to recommend safe throttle and steering actions to the rider. Our main contributions are:

\begin{itemize}

\item A novel perception-driven ARAS pipeline for motorcycles that proactively identifies scene-level risks relevant to two-wheeled vehicles.

\item A spatial risk representation that fuses VLM-based semantic reasoning with prompt-conditioned segmentation to produce dense, per-pixel hazard costs suitable for planning.

\item An adapted sampling-based local planner to model motorcycle-specific dynamics and integrates our risk map to generate risk-aware trajectories.



\end{itemize}


\section{Related work}

In this section, we review prior work on advanced driver and rider assistance systems, hazard detection methods, and the use of large models for autonomous vehicles.

\subsection{Advanced Driver \& Rider Assistance Systems}

Advanced Driver Assistance Systems (ADAS) have achieved significant success in automotive applications, with features like lane-keeping, adaptive cruise control, and blind-spot monitoring \cite{ayyasamy2022comprehensive, ziebinski2017review}.  However, Advanced Rider Assistance Systems (ARAS) for motorcycles remain relatively underexplored due to the unique instability and dynamics of two-wheeled vehicles \cite{ait2024effect}. Early work explored ABS and traction control \cite{rizzi2015effectiveness}, but more recent systems such as BMW Motorrad’s ARAS incorporate radar-based collision warnings and adaptive braking \cite{hamm2022bmw}. These systems rely on a robust perception of the vehicle’s surroundings to identify hazards and assist the human rider \cite{lot2011advanced,mavsek2022design,kuschefski2011advanced}.
Despite progress, ARAS systems lack contextual understanding of the environment, which is critical for the rider's safety.

\subsection{Hazard Detection}

Detecting and responding to unexpected road hazards is critical for both autonomous vehicles and human riders. Conventional perception methods excel at recognizing common classes of objects (e.g., cars, pedestrians) using detection and segmentation methods \cite{mao20233d,liang2022edge,chen2023edge}. However, they often fail to detect unusual or out-of-distribution hazards on the road \cite{yang2023uncertainties}. To bridge this gap, researchers have explored open-set hazard classification and anomaly detection techniques for autonomous driving \cite{bogdoll2022anomaly}.  For example, Armstrong developed a vision-based system to flag traffic anomalies beyond the standard object categories \cite{aboah2021vision}. Recent work also leverages multimodal cues and vision-language models to improve hazard understanding; Chen et al. integrate semantic context with visual images to better identify “edge case” dangers that traditional methods did not detect \cite{chen2025insight}. These approaches primarily focus on detection without integrating hazard reasoning into downstream planning or rider assistance.

\subsection{Large models for Autonomous Vehicles}

Large foundation models have recently been applied to autonomous driving tasks to provide high-level reasoning and scene understanding \cite{cui2024survey}. Vision-Language Models (VLMs) and Large Language Models (LLMs) trained on vast data have shown improvements on AV perception and planning \cite{fu2024drive, cui2024receive, cui2023drivellm}. For instance, Mao et al. introduce GPT-Driver, which uses a GPT-3.5 model to generate motion plans and explain decisions in driving scenarios \cite{mao2023gpt}. Similarly, Li et al. propose LLaDA, which leverages an LLM to interpret regional traffic rules and adapt driving policies accordingly, enabling vehicles to handle diverse environments in a zero-shot manner \cite{li2024driving}. These approaches demonstrate the potential of large models to enhance scene understanding in autonomous driving. However, these advances have yet to be applied to ARAS, particularly for simulated and real-world motorcycle navigation.

\section{Background}
In this section, we explain the symbols and our problem formulation. 


\subsection{Symbols and Notations}
We highlight the symbols of images and cost maps we use in Table (\ref{tab:symbols}).

\begin{table}[h]
\caption{\justifying \small Key symbols used in our ARAS. Cost maps and masks are $H \times W$ and aligned with $I_{\text{RGB}}$. 
} 
\centering
\small
\begin{tabular}{|c|l|}
\hline
\textbf{Symbol} & \textbf{Definition} \\
\hline
$ \mathbf{s} $ & Motorcycle state: velocity $v$ and lean angle $\phi$. \\
\hline
$ I_{\text{RGB}} $ & Input front camera image ($H \times W \times 3$). \\
\hline
$ \mathcal{H} $ & Hazards identified by the VLM. \\
\hline
$ \mathcal{M}_i $ & Binary segmentation mask for hazard $h_i$. \\
\hline
$ \mathcal{C}_i(m,n) $ & Risk cost for hazard $h_i$ at pixel $(m,n)$. \\
\hline
$ \mathcal{C}_{\text{risk}} $ & Final risk map encoding per-pixel hazard severity. \\
\hline
$ \mathbf{x}_t $ & Motorcycle state: $[x, y, \theta, v, \delta]$. \\
\hline
$ \text{traj} $ & Simulated trajectory from current state. \\
\hline
$ \mathcal{U} $ & Discrete set of candidate control inputs. \\
\hline
$ J $ & Cost function used to evaluate each trajectory. \\
\hline
\end{tabular}
\label{tab:symbols}
\vspace{-10pt}
\end{table}

\begin{figure*}[t]
\centering
\includegraphics[width=\textwidth, height=11cm, keepaspectratio]{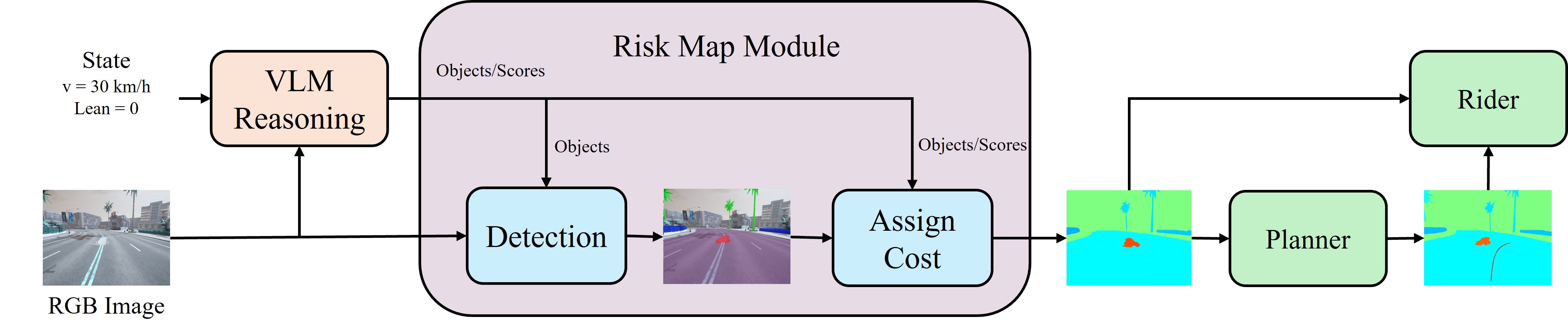} 
\caption{\justifying Overall architecture of our system: The VLM first receives the front-facing RGB frame along with the motorcycle state and performs chain-of-thought scene reasoning. It identifies relevant objects (e.g., “tree”, “road”, “pothole”, “grass”) and assigns contextual risk scores, for example (0.2, 0, 0.7, 0.1), reflecting their relative hazard to the rider. These identified objects and scores are then passed to an open-vocabulary segmentation model, which spatially localizes each hazard in the image. The Risk Map Module combines the VLM’s contextual scores with geometric, depth-based, and detection-confidence factors to produce a dense per-pixel risk map. A sampling-based planner evaluates candidate trajectories using this map and recommends risk-aware control actions to the rider. This architecture integrates high-level VLM reasoning with precise spatial localization to support hazard-aware motorcycle guidance.}
\label{fig:arch}
\end{figure*}

\subsection{Problem formulation}

Given RGB images from a front-facing RGB camera and the current motorcycle state $\mathbf{s}_t = [v_t, \phi_t]$, where $v_t$ is the velocity and $\phi_t$ is the lean angle at time $t$, the goal of the ARAS is to detect and localize road hazards $\mathcal{H}_t$ (e.g., potholes, water puddles), reason about their potential risk to the rider based on both visual context and bike's state, and generate a spatial risk representation $\mathcal{R}_t$ of the environment. 
This risk map is used by a sampling-based planner to compute an appropriate control actions $\mathbf{u}_t $ 
to ensure safe navigation.
\section{Methodology}

In this section, we outline the methodology of our approach. The pipeline of our system is shown in Figure \ref{fig:arch} and the algorithm is summarized in Algorithm 1.

\subsection{Scene Understanding} \label{sec:reasoning}

Our ARAS pipeline begins by leveraging a large Vision-Language Model (VLM) to jointly analyze each front-facing RGB frame and the motorcycle's current state, $\mathbf{s} = [v, \phi]$. The VLM is queried using a structured chain-of-thought (CoT) prompt to reason about the scene and identify objects of interest that may pose safety-critical risks to the rider (e.g., potholes, water puddles, loose gravel). For each identified object, the VLM infers a contextual risk score by considering both the visual semantics and the bike's current physical state. We query the VLM only once per scene and assume that the set of hazards remains unchanged during this time.
This reasoning process yields a structured output comprising a set of potential hazard objects, $\mathcal{H}_t = \{h_1, h_2, ..., h_n\}$, and their corresponding contextual risk scores, $\mathcal{C}_t = \{c_1, c_2, ..., c_n\}$, where each $c_i = c_i^{\text{vlm}} \in [0, 1]$ reflects the relative severity of object $h_i$ with respect to rider safety.

\subsection{Risk Map Construction}

Following the reasoning stage, we construct a spatial risk map, $\mathcal{C}_{\text{risk}}$, that encodes the severity of hazards identified by the VLM. This map integrates grounded object segmentation with multiple contributing factors to represent hazard-specific risk distributions across the scene.

\subsubsection{Segmentation of Hazards}

Given the set of objects of interest $\mathcal{H}_t = \{h_1, h_2, \dots, h_n\}$ identified by the VLM, we estimate their spatial extents in the RGB image $I_{\text{RGB}} \in \mathbb{R}^{H \times W \times 3}$ using \cite{ren2024grounded}. For each object $h_i \in \mathcal{H}_t$, the image and its corresponding text prompt are jointly processed to generate a binary segmentation mask $\mathcal{M}_i \in \{0,1\}^{H \times W}$, where each pixel $(m,n)$ denotes whether it belongs to the region associated with hazard $h_i$. This produces prompt-conditioned masks that localize each hazard.

\subsubsection{Hazard Risk Scoring Based on Multi-Factor Cost}






To estimate the relative severity of each segmented hazard, we assign a per-pixel risk cost $\mathcal{C}_i(m,n)$ to all pixels $(m,n)$ within the corresponding mask $\mathcal{M}_i$. The total hazard cost combines four complementary factors: contextual, area-based, confidence-based, and depth-based terms as:

\begin{equation}
    \mathcal{C}_i(m, n) = 
    \alpha_{\text{vlm}} \, c^{\text{vlm}}_i +
    \alpha_{\text{area}} \, c^{\text{area}}_i +
    \alpha_{\text{conf}} \, c^{\text{conf}}_i +
    \alpha_{\text{depth}} \, c^{\text{depth}}_i,
    \label{eq:total_cost}
\end{equation}

where $\alpha_{\text{vlm}}, \alpha_{\text{area}}, \alpha_{\text{conf}}, \alpha_{\text{depth}} \in \mathbb{R}$ are adjustable weight coefficients.

\noindent\textbf{Contextual Risk ($c^{\text{vlm}}_i$):}  
This term represents the contextual risk assigned by the VLM based on semantic reasoning of the scene, the current motorcycle state $(s_t)$, and the textual prompt $T_{\text{prompt}}$.  
\begin{equation}
    c^{\text{vlm}}_i = \text{VLM}(I_{\text{RGB}}, s_t, T_{\text{prompt}}), \quad
    c^{\text{vlm}}_i \in [0, 1].
\end{equation}

\noindent\textbf{Area Score ($c^{\text{area}}_i$):}  
A sigmoid-based function captures the normalized area ratio of the segmented hazard:
\begin{equation}
    c^{\text{area}}_i = 
    \sigma \left( \kappa_{\text{a}} \left( 
    \frac{|\mathcal{M}_i|}{H \times W} - \eta_{\text{a}} 
    \right) \right),
\end{equation}
where $\sigma(\cdot)$ is the sigmoid function, $|\mathcal{M}_i|$ is the pixel area of the mask, and $\kappa_{\text{a}}, \eta_{\text{a}}$ are scale and offset parameters.

\noindent\textbf{Confidence Score ($c^{\text{conf}}_i$):}  
Derived from detection confidence $p_i$ of the grounding model \cite{ren2024grounded}.

\noindent\textbf{Depth Score ($c^{\text{depth}}_i$):}  
For depth-sensitive hazards (e.g., potholes), we estimate the maximum depth deviation within the segmented region and normalize it relative to a reference depth $d_{\text{ref}}$:
\begin{equation}
    c^{\text{depth}}_i = 
    \min\left(1, \frac{d_i}{d_{\text{ref}}}\right),
\end{equation}
where $d_i$ is the estimated depth difference between the fitted road plane and the pothole region.

\subsubsection{Pixel-wise Risk Evaluation}

In cases where multiple hazard masks overlap at a given pixel, we prioritize the highest severity by selecting the maximum cost among all hazard-specific maps. The final risk map $\mathcal{C}_{\text{risk}} \in [0, C_{\max}]^{H \times W}$ is defined as:

\begin{equation}
\mathcal{C}_{\text{risk}}(m, n) = \max_{i=1}^{j} \mathcal{C}_i(m, n),
\end{equation}

where $j$ is the number of hazards in $\mathcal{H}_t$. This operation ensures that, for each spatial location, the most critical hazard is represented in the final cost map. The resulting $\mathcal{C}_{\text{risk}}$ encodes a dense, spatial distribution of context-aware hazard severity across the scene. 

\begin{algorithm}[h]
\caption{VLM-Based Advanced Rider Assistance System}
\label{alg:aras}
\begin{algorithmic}[1]
\REQUIRE RGB frame $I_{\text{RGB},t}$, state $s_t=[v_t,\phi_t]$, prompt $T_{\text{prompt}}$, goal $g$, intrinsics $K$, extrinsics $\mathbf{T}_{cw}$
\ENSURE Risk map $\mathcal{C}_{\text{risk}}$, control $(a^*, \dot{\delta}^*)$
\STATE Query VLM$(I_{\text{RGB},t}, s_t, T_{\text{prompt}})$ to obtain hazards $\mathcal{H}_t=\{h_i\}$ and contextual scores $c^{\text{vlm}}_i$
\FOR{$h_i \in \mathcal{H}_t$}
    \STATE Obtain mask $\mathcal{M}_i$
    \STATE Compute $c^{\text{area}}_i$, $c^{\text{conf}}_i$, $c^{\text{depth}}_i$
    \STATE $\mathcal{C}_i(m,n) \!\leftarrow\! \alpha_{\text{vlm}}c^{\text{vlm}}_i + \alpha_{\text{area}}c^{\text{area}}_i + \alpha_{\text{conf}}c^{\text{conf}}_i + \alpha_{\text{depth}}c^{\text{depth}}_i$
\ENDFOR
\STATE $\mathcal{C}_{\text{risk}}(m,n) \!\leftarrow\! \max_i \mathcal{C}_i(m,n)$
\STATE $\psi_{\text{risk}} \leftarrow \mathcal{C}_{\text{risk}}(m,n), K, \mathbf{T}_{cw}$
\FOR{$(a,\dot{\delta}) \in \mathcal{U}$}
    \STATE $J(q) \!\leftarrow\! \beta_{\text{goal}}\psi_{\text{goal}} + \beta_{\text{speed}}\psi_{\text{speed}} + \beta_{\text{risk}}\psi_{\text{risk}}$
\ENDFOR
\STATE $(a^*, \dot{\delta}^*) \!\leftarrow\! \arg\min_{(a,\dot{\delta})\in\mathcal{U}} J(q)$
\RETURN $\mathcal{C}_{\text{risk}}$, $(a^*, \dot{\delta}^*)$
\end{algorithmic}
\end{algorithm}

\subsection{Risk-Aware Motion Planning}

To support real-time hazard-aware decision-making, we implement a sampling-based planner that recommends optimal control actions based on the current vehicle state and the risk map $\mathcal{C}_{\text{risk}}$. The planner is based on the dynamic window approach (DWA) \cite{dwa}, 

Let $\mathbf{x}_t = [x, y, \theta, v, \delta]$ denote the current state of the motorcycle, where $(x, y)$ is the position, $\theta$ is the heading, $v$ is the velocity, and $\delta$ is the steering angle. Each control pair $(a, \dot{\delta}) \in \mathcal{U}$ is forward simulated using bicycle dynamics over a prediction horizon $T$:

\begin{align}
\dot{x} &= v \cos(\theta) \\
\dot{y} &= v \sin(\theta) \\
\dot{\theta} &= \frac{v}{L} \tan(\delta) \\
\dot{v} &= a 
\end{align}

Each rollout produces a trajectory $q$, which is evaluated using the following cost function:

\begin{equation}
J(q) = 
\beta_{\text{goal}} \cdot \psi_{\text{goal}}(q) + 
\beta_{\text{speed}} \cdot \psi_{\text{speed}}(q) + 
\beta_{\text{risk}} \cdot \psi_{\text{risk}}(q)
\end{equation}


The terms are defined as follows:
\begin{itemize}
    \item $\psi_{\text{goal}}(q)$: Euclidean distance from the final state in $q$ to the goal.
    \item $\psi_{\text{speed}}(q)$: Encourages higher linear (translational) velocity at the trajectory endpoint, computed as a penalty on low terminal speed, $v_{\max} - v_t$, where $v_{\max}$ is a user-defined upper bound on the desired forward velocity.
    \item $\psi_{\text{risk}}(q)$: Average cost sampled from $\mathcal{C}_{\text{risk}}$ along $q$, projected into the image frame using the current camera transform.
\end{itemize}

\begin{figure*}[t]
\centering
\includegraphics[width=\textwidth, height=11cm, keepaspectratio]{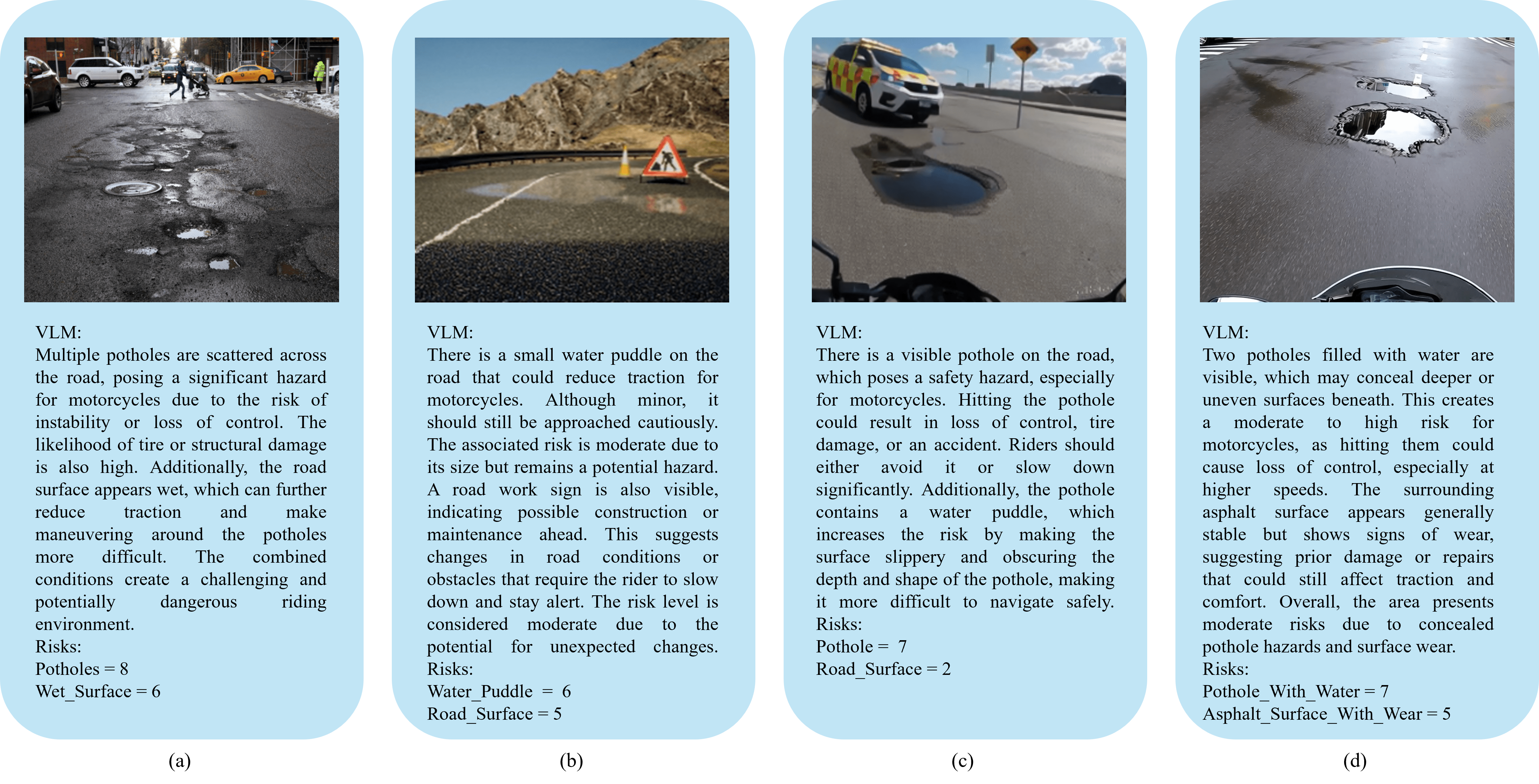}
\caption{\justifying
Zero-shot VLM reasoning across diverse road conditions. (a) A real-world road with multiple potholes and poor surface quality, where the VLM assesses overall surface condition and contextual hazard severity. (b) A CARLA simulation with a cone, where the VLM recognizes a puddle and a road-work sign indicating moderate contextual risk. (c) A synthetic scene where the VLM distinguishes a pothole beneath a water puddle and assigns higher risk to the pothole. (d) A pothole filled with water on a good road, where the VLM differentiates between a simple puddle and concealed surface damage.}
\label{fig:vlm_samples_traj}
\end{figure*}

\subsubsection{Trajectory Projection into Image Space}

To evaluate trajectory risk with respect to the spatial cost map $\mathcal{C}_{\text{risk}}$, we project candidate trajectory waypoints from world coordinates into the image plane of the onboard camera.

Let each 3D point $\mathbf{p}_w = (x_w, y_w, z_w)$ represent a position in the world frame, corresponding to a simulated waypoint along a candidate trajectory. Given the camera extrinsics in the form of a transformation matrix $\mathbf{T}_{cw}$ that maps world coordinates to the camera frame, the point in the camera frame is:

\begin{equation}
\mathbf{p}_c = \mathbf{T}_{cw} \cdot \mathbf{p}_w.
\end{equation}

Given the camera intrinsics matrix $K \in \mathbb{R}^{3 \times 3}$, we project the point into the image plane using the pinhole model:

\begin{equation}
\mathbf{p}_i = K \cdot \mathbf{p}_c, \quad
(m, n) = \left( \frac{p_{i,x}}{p_{i,z}}, \frac{p_{i,y}}{p_{i,z}} \right)\end{equation}

where $(m, n)$ denotes the pixel coordinates in the image. The risk cost at each trajectory point is obtained by sampling $\mathcal{C}_{\text{risk}}(m, n)$.

Having defined how risk is evaluated along projected trajectories, we now search for optimal actions by discretizing the feasible control space $\mathcal{U}$ by sampling acceleration $a \in [a_{\min}, a_{\max}]$ and steering rate $\dot{\delta} \in [\dot{\delta}_{\min}, \dot{\delta}_{\max}]$. For each sampled pair, we simulate its resulting trajectory $q$ using the dynamics above and score it using $J(q)$. The best actions are then selected via:

\begin{equation}
(a^*, \dot{\delta}^*) = \arg\min_{(a, \dot{\delta}) \in \mathcal{U}} J(q)
\end{equation}


\section{Experiments and Results}
We evaluate our proposed ARAS through experiments designed to examine its two core functions: (1) providing interpretable,
scene-level risk maps that a rider could use as hazard feedback, and (2) recommending safe, risk-aware trajectories that a rider
could follow. 
We conduct qualitative evaluations using both synthetic video scenarios and the 
CARLA simulator, and quantitative evaluations using the Kawasaki Ninja motorcycle model in CARLA 0.9.14 with GPT-4o VLM for reasoning. We implement our system using a laptop with an NVIDIA RTX 4090 GPU.
\subsection{Qualitative Evaluation of Hazard Interpretation}

We first evaluate our proposed ARAS pipeline qualitatively to illustrate how the Vision-Language Model (VLM) interprets hazards and how these interpretations are integrated into risk maps and downstream planning.

\subsubsection{VLM Reasoning}

Figure \ref{fig:vlm_samples_traj} illustrates zero-shot VLM reasoning across diverse road conditions, that produces hazard scores on a 0–10 scale (0 = no risk, 10 = high risk). In (a), the VLM evaluates the overall road conditions and identifies multiple potholes and degraded asphalt surfaces as high-risk regions due to poor traction and uneven geometry. In (b) and (c), it interprets contextual cues such as cones, roadwork signs, and service vehicles, inferring that these elements imply construction zones or temporary hazards. In (c) and (d), the VLM differentiates between visually similar but semantically distinct cases, a surface water puddle versus a pothole filled with water, and assigns different risk levels based on their potential impact on motorcycle stability. 
These examples show how the VLM enhances contextual scene understanding by combining road condition analysis, environmental cues, and surface semantics to support hazard-aware motorcycle assistance.

\subsubsection{Interpretable Risk Maps}

\begin{figure}[!t]
\centering
\includegraphics[width = \columnwidth,  height=0.795\textheight]{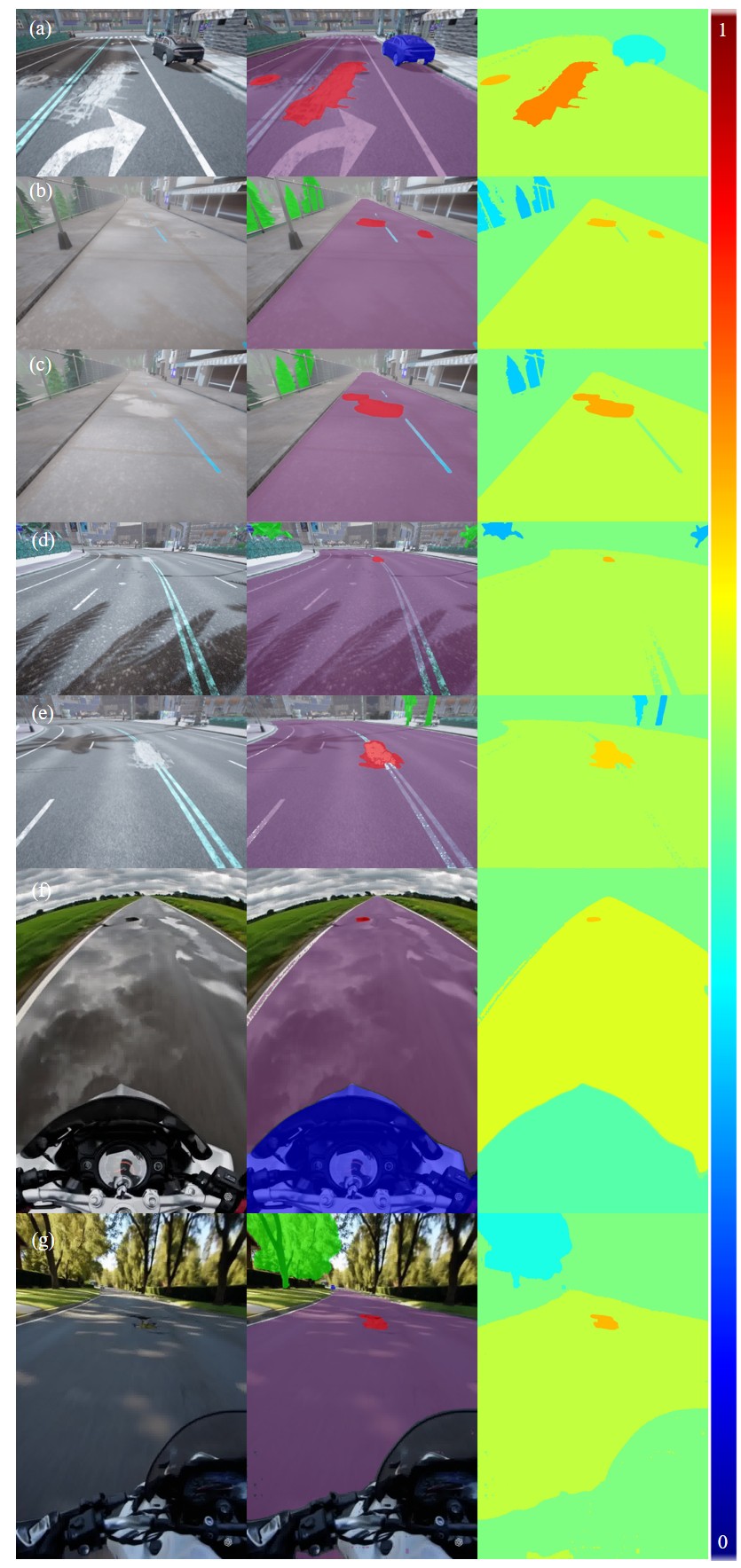} 
\caption{\justifying Risk map samples from our method using simulated and synthetic scenes. (a–e) show simulated scenarios in CARLA, and
(f–g) show synthetic scenes. The evaluated hazards include potholes in (a), (d), (e), and (g), and water puddles in (b), (c), and
(f). Left: RGB image; center: segmentation output; right: generated risk map. The risk map integrates contextual, area, depth,
and detection confidence factors. The color bar indicates normalized risk intensity, where 1 represents the highest risk and 0 the
lowest.}
\label{fig:risk_maps_samples}
\end{figure}

Building on the VLM’s contextual scoring, our ARAS fuses its outputs with segmentation-based attributes (depth and area of the hazard) to produce dense, per-pixel risk maps. Figure \ref{fig:risk_maps_samples} illustrates scenarios with hazards such as potholes and water puddles, where our method concentrates high costs around hazard regions, yielding interpretable risk maps that highlight safety-critical areas. Compared to raw VLM scores, these spatial maps provide a good representation for planning.



\begin{figure}[t]
\centering
\includegraphics[width = 0.8\columnwidth]{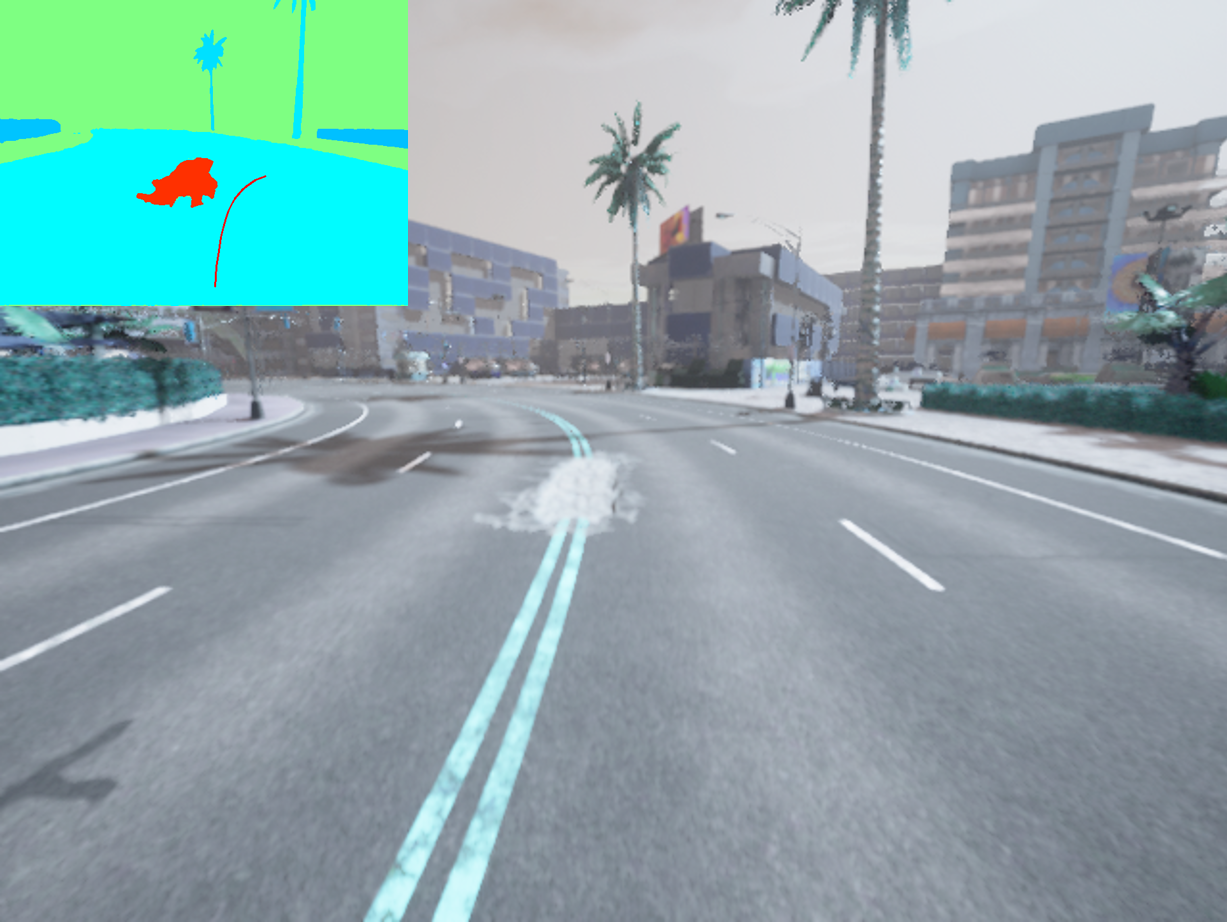} 
\caption{ \justifying Trajectory generation in CARLA. The motorcycle approaches a pothole, and our method generates a risk map that guides the planner to recommend a safe trajectory that deviates around the hazard while maintaining progress toward the goal.}
\label{fig:carla_traj}
\end{figure}

\subsection{Evaluation of Recommended Trajectories}

To assess the quality of the trajectories our ARAS system would recommend to a rider, we evaluate the full pipeline in simulation. Figure~\ref{fig:carla_traj} illustrates a representative example: after the VLM assigns a semantic risk score to the identified hazard, this score is grounded into a spatial risk map through segmentation, and the sampling-based planner uses the resulting map to generate a path that deviates safely around the pothole while maintaining forward progress. 

We further evaluate these suggested trajectories quantitatively using three simulated scenarios.

\subsubsection{Comparison methods \& Metrics}

We compare our approach against baseline planner \cite{dwa} and an ablation of our method. 
The baseline planner is a sampling-based controller that relies on goal distance, obstacles, and vehicle dynamics for trajectory generation; and 
The ablated version excludes the VLM contextual cost, where the risk map includes only geometric and detection-based attributes without semantic reasoning.

We evaluate navigation performance using two evaluation metrics:
\begin{itemize}

   \item  \textbf{Success Rate:} The percentage of successful trials in which the motorcycle body does not intersect any hazard region. A higher value indicates improved hazard-avoidance behavior.

   \item \textbf{Hazard Exposure Distance:} The minimum distance (meters) between the motorcycle’s trajectory and the nearest hazard centroid throughout the run. Smaller values indicate higher exposure risk, while larger distances reflect safer clearance from hazards.
\end{itemize}

\subsubsection{CARLA Evaluation Scenarios}

We evaluate our method across the following simulated scenarios:

\begin{itemize}
    \item \textbf{Scenario 1:} A small pothole positioned in the center of the motorcycle’s lane.
    \item \textbf{Scenario 2:} A large pothole positioned at the center of the road.
    \item \textbf{Scenario 3:} A large pothole in the center of the motorcycle’s lane with a cone warning sign.
\end{itemize}

\begin{figure*}[t]
\centering
\includegraphics[width=\textwidth]{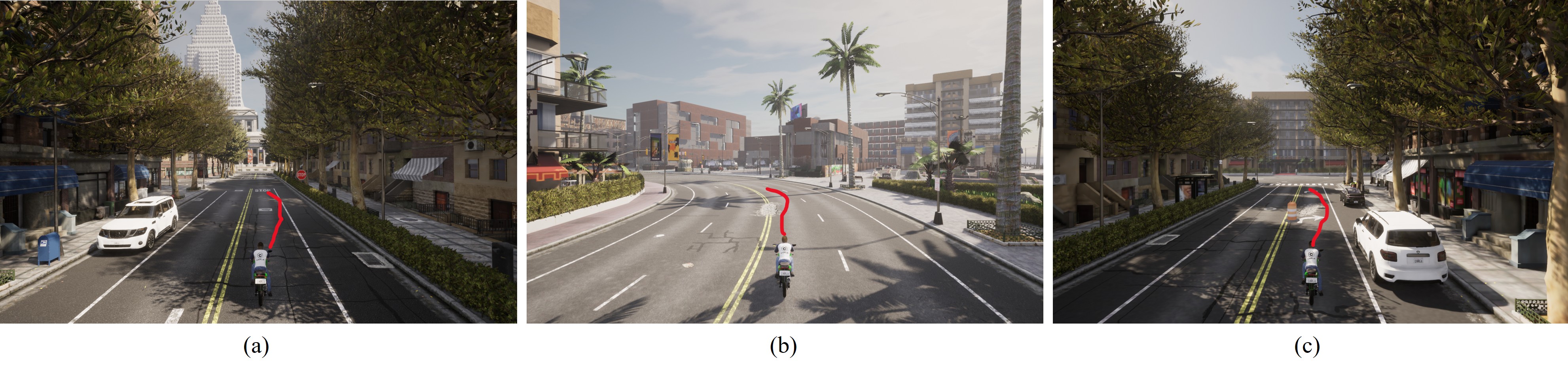}
\caption{\justifying
Motorcycle navigation in CARLA showing trajectories generated by our method (red). (a) Scenario 1, a small pothole in the motorcycle’s lane, (b) Scenario 2, a large pothole centered on the road, and (c) Scenario 3, a large pothole with a nearby cone warning sign.
}
\label{fig:carla_expirments}
\end{figure*}


\subsubsection{Discussion}
We evaluate our system qualitatively in Figure \ref{fig:carla_expirments} and quantitatively in Table~\ref{tab:navigation-comparison} to assess how well the recommended trajectories help avoid hazards. Across all scenarios, our method suggests safer and more consistent paths than the baseline planner. Qualitatively, our risk-aware planner redirects the motorcycle around potholes and warning signs (cones), whereas the baseline frequently
passes over or very near the hazard. Quantitatively, our method achieves higher success rates and larger hazard–clearance distances, which indicates improved hazard avoidance and safer trajectory behavior. Removing the VLM contextual cost reduces performance, which highlights the importance of semantic reasoning for interpreting scene-level hazard cues. Overall, these findings demonstrate that integrating VLM-based reasoning with spatial risk mapping enhances the motorcycle's ability to anticipate and avoid surface hazards.

\begin{table}
\caption{\small \justifying Quantitative performance comparison for three scenarios over 50 trials. The scenarios include potholes of different sizes and shapes.}
\centering
\resizebox{0.9\columnwidth}{!}{
\begin{tabular}{c c c c c} 
\toprule
\textbf{Metrics} & \textbf{Method} & \textbf{Scn. 1} & \textbf{Scn. 2} & \textbf{Scn. 3} \\  
\midrule

\multirow{3}{*}{\makecell{\textbf{Success}\\ \textbf{ Rate (\%) $\uparrow$}}}
 & DWA Planner \cite{dwa} & -    & -    & -    \\
 & w/o VLM                & 74 & 62 & 52 \\
 & Ours                   & 78 & 70 & 68 \\
\midrule

\multirow{3}{*}{\makecell{\textbf{Hazard}\\ \textbf{ Exposure}\\ \textbf{ Distance}}}
 & DWA Planner \cite{dwa} & 0.22 & 0.33 & 0.19 \\
 & w/o VLM                & 0.31 & 0.39 & 0.35 \\
 & Ours                   & 0.32 & 0.45 & 0.38 \\
\bottomrule
\end{tabular}
}
\label{tab:navigation-comparison}
\vspace{-20pt}
\end{table}


\section{Conclusions, Limitations and Future Work}

We presented an Advanced Rider Assistance System (ARAS) that integrates the reasoning of vision-language models with spatial localization from segmentation models to construct dense, hazard-aware risk maps for motorcycles. These maps capture both contextual and geometric hazard attributes, which enable a sampling-based planner tailored to motorcycle dynamics to generate safe and risk-aware control actions. Our experiments in the CARLA simulator demonstrate that this hybrid pipeline improves success rates and reduces exposure to hazards compared to the baseline method. In addition, we qualitatively evaluated the risk maps in synthetic videos, which showed interpretable hazard localization and risk-aware motion guidance. While the lean angle is incorporated in the VLM input to support the context-aware risk reasoning, it is not modeled in the planner’s dynamics. This simplification may limit accuracy under high-speed or aggressive maneuvers, which we plan to address in future work. 
Another limitation of our system is the latency of the VLM reasoning. Because inference is slow, we query the VLM only once per scene and assume hazards remain unchanged during the short planning horizon. This may limit performance in more complex settings. In future work, we plan to distill or fine-tune a smaller VLM that can run locally and update the contextual score more frequently.
Ultimately, we hope this work motivates further research in ARAS and highlights the need for motorcycle datasets and realistic two-wheeled simulation platforms to advance the field.



\bibliographystyle{IEEEtran}
\bibliography{refs}






\end{document}